\documentclass[runningheads]{llncs}
\usepackage{graphicx}
\usepackage{amsfonts}
\usepackage{amsmath}
\usepackage{makecell}
\usepackage{booktabs}
\usepackage{multirow}
\usepackage{xcolor}
\usepackage{soul}
\usepackage{color}
\usepackage{marvosym}
\usepackage{caption}

\makeatletter
\newcommand{\printfnsymbol}[1]{%
  \textsuperscript{\@fnsymbol{#1}}%
}
\makeatother
\begin{document}
\title{TransFuse: Fusing Transformers and CNNs for Medical Image Segmentation} 
%
%

\author{Yundong Zhang\thanks{These authors contributed equally to this work.} \inst{1} \and
Huiye Liu\printfnsymbol{1} \inst{1,2} $^{\textrm{(\Letter)}}$ \and
Qiang Hu\inst{1}}

\authorrunning{Y. Zhang et al.}
\institute{Rayicer, Suzhou, China\\
\email{{huiyeliu}@rayicer.com}\\
\and
Georgia Institute of Technology, Atlanta, GA, USA\\
}

\maketitle              
\begin{abstract}
Medical image segmentation - the prerequisite of numerous clinical needs - has been significantly prospered by recent advances in convolutional neural networks (CNNs). However, it exhibits general limitations on modeling explicit long-range relation, and existing cures, resorting to building deep encoders along with aggressive downsampling operations, leads to redundant deepened networks and loss of localized details. Hence, the segmentation task awaits a better solution to improve the efficiency of modeling global contexts while maintaining a strong grasp of low-level details. 
In this paper, we propose a novel parallel-in-branch architecture, TransFuse, to address this challenge. TransFuse combines Transformers and CNNs in a parallel style, where both global dependency and low-level spatial details can be efficiently captured in a much shallower manner. Besides, a novel fusion technique - BiFusion module is created to efficiently fuse the multi-level features from both branches. 
Extensive experiments demonstrate that TransFuse achieves the newest state-of-the-art results on both 2D and 3D medical image sets including polyp, skin lesion, hip, and prostate segmentation, with significant parameter decrease and inference speed improvement.

\end{abstract}
\section{Introduction}

Convolutional neural networks (CNNs) have attained unparalleled performance in numerous medical image segmentation tasks~\cite{hesamian2019deep,isensee2019automated},  
such as multi-organ segmentation, liver lesion segmentation, brain 3D MRI, etc., as it is proved to be powerful at building hierarchical task-specific feature representation by training the networks end-to-end. 
Despite the immense success of CNN-based methodologies, its lack of efficiency in capturing global context information remains a challenge. The chance of sensing global information is equaled by the risk of efficiency, because existing works obtain global information by generating very large receptive fields, which requires consecutively down-sampling and stacking convolutional layers until deep enough. This brings several drawbacks: 1) training of very deep nets is affected by the diminishing feature reuse problem~\cite{srivastava2015highway}, where low-level features are washed out by consecutive multiplications; 2) local information crucial to dense prediction tasks, e.g., pixel-wise segmentation, is discarded, as the spatial resolution is reduced gradually; 3) training parameter-heavy deep nets with small medical image datasets tends to be unstable and easily overfitting. 
Some studies~\cite{wang2018non} use the non-local self-attention mechanism to model global context; however, the computational complexity of these modules typically grows quadratically with respect to spatial size, thus they may only be appropriately applied to low-resolution maps.

Transformer, originally used to model sequence-to-sequence predictions in NLP tasks~\cite{vaswani2017attention}, has recently attracted tremendous interests in the computer vision community. The first purely self-attention based vision transformers (ViT) for image recognition is proposed in~\cite{dosovitskiy2020image}, which obtained competitive results on ImageNet~\cite{deng2009imagenet} with the prerequisite of being pretrained on a large external dataset. 
SETR~\cite{zheng2020rethinking} replaces the encoders with transformers in the conventional encoder-decoder based networks to successfully achieve state-of-the-art (SOTA) results on the natural image segmentation task. While Transformer is good at modeling global context, it shows limitations in capturing fine-grained details, especially for medical images. We independently find that SETR-like pure transformer-based segmentation network produces unsatisfactory performance, due to lack of spatial inductive-bias in modelling local information (also reported in~\cite{chen2021transunet}).

To enjoy the benefit of both, efforts have been made on combining CNNs with Transformers, e.g., TransUnet~\cite{chen2021transunet}, which first utilizes CNNs to extract low-level features and then passed through transformers to model global interaction. With skip-connection incorporated, TransUnet sets new records in the CT multi-organ segmentation task. However, past works mainly focus on replacing convolution with transformer layers or stacking the two in a sequential manner. To further unleash the power of CNNs plus Transformers in medical image segmentation, in this paper, we propose a different architecture---\textit{TransFuse}, which runs shallow CNN-based encoder and transformer-based segmentation network in parallel, followed by our proposed \textit{BiFusion} module where features from the two branches are fused together to jointly make predictions. TransFuse possesses several advantages: 1) both low-level spatial features and high-level semantic context can be effectively captured; 2) it does not require very deep nets, which alleviates gradient vanishing and feature diminishing reuse problems; 3) it largely improves efficiency on model sizes and inference speed, enabling the deployment at not only cloud but also edge. To the best of our knowledge, TransFuse is the first parallel-in-branch model synthesizing CNN and Transformer. Experiments demonstrate the superior performance against other competing SOTA works.

\section{Proposed Method}
\begin{figure}
\centering
\includegraphics[width=\linewidth]{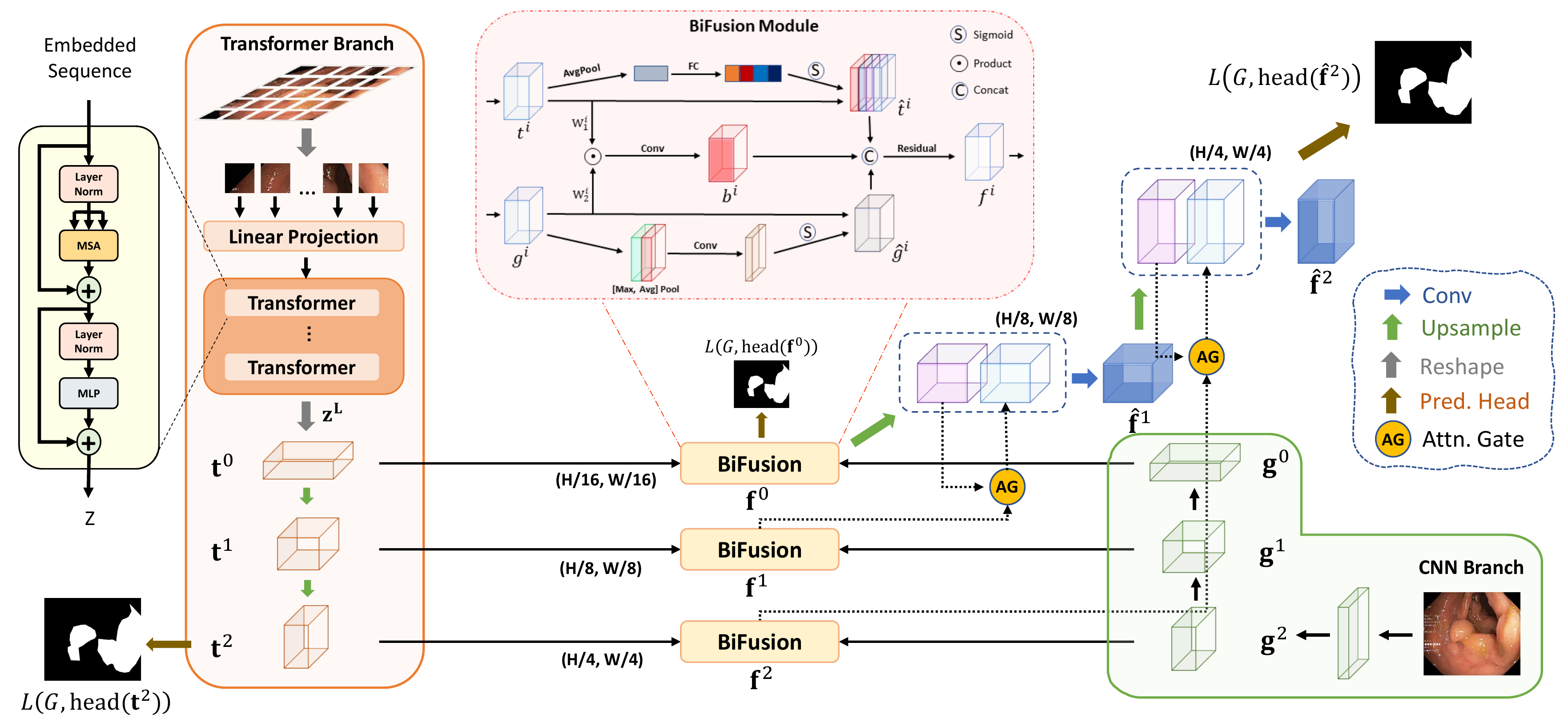}
\caption{Overview of TransFuse (best viewed in color): two parallel branches - CNN (bottom right) and transformer (left) fused by our proposed BiFusion module.} 
\label{fig1}
\end{figure}

As shown in Fig. \ref{fig1}, TransFuse consists of two parallel branches processing information differently: 1) CNN branch, which gradually increases the receptive field and encodes features from local to global; 2) Transformer branch, where it starts with global self-attention and recovers the local details at the end. Features with same resolution extracted from both branches are fed into our proposed BiFusion Module, where self-attention and bilinear Hadamard product are applied to selectively fuse the information. Then, the multi-level fused feature maps are combined to generate the segmentation using gated skip-connection \cite{schlemper2019attention}.
There are two main benefits of the proposed branch-in-parallel approach: firstly, 
by leveraging the merits of CNNs and Transformers, we argue that TransFuse can capture global information without building very deep nets while preserving sensitivity on low-level context; secondly, our proposed BiFusion module may simultaneously exploit different characteristics of CNNs and Transformers during feature extraction, thus making the fused representation powerful and compact.

\textit{\textbf{Transformer Branch.}}
The design of Transformer branch follows the typical encoder-decoder architecture. Specifically, the input image $\mathbf{x}\in\mathbb{R}^{H\times W\times 3}$ is first evenly divided into $N=\frac{H}{S}\times\frac{W}{S}$ patches, where $S$ is typically set to 16. The patches are then flattened and passed into a linear embedding layer with output dimension $D_0$, obtaining the raw embedding sequence $\mathbf{e}\in\mathbb{R}^{N\times D_0}$. To utilize the spatial prior, a learnable positional embeddings of the same demension is added to $\mathbf{e}$. The resulting embeddings $\mathbf{z}^{0}\in\mathbb{R}^{N\times D_0}$ is the input to Transformer encoder, which contains $L$ layers of multiheaded self-attention (MSA) and Multi-layer Perceptron (MLP). We highlight that the self-attention (SA) mechanism, which is the core principal of Transformer, updates the states of each embedded patch by aggregating information globally in every layer:
\begin{equation}\label{Eq1}
    \textup{SA}(\mathbf{z}_{i}) =  \textup{softmax}\left(\frac{\mathbf{q_i}\mathbf{k}^{T}}{\sqrt{D_h}}\right)\mathbf{v},
\end{equation}
where $[\mathbf{q, k, v}]=\mathbf{z}\mathbf{W}_{qkv}$, $\mathbf{W}_{qkv}\in\mathbb{R}^{D_0\times 3D_h}$ is the projection matrix and vector $\mathbf{z}_i\in\mathbb{R}^{1\times D_0},\mathbf{q_i}\in\mathbb{R}^{1\times D_h}$ are the $i^{th}$ row of $\mathbf{z}$ and $\mathbf{q}$, respectively. MSA is an extension of SA that concatenates multiple SAs and projects the latent dimension back to $\mathbb{R}^{D_0}$, and MLP is a stack of dense layers (refer to~\cite{dosovitskiy2020image} for details of MSA and MLP). 
Layer normalization is applied to the output of the last transformer layer to obtain the encoded sequence $\mathbf{z}^L\in\mathbb{R}^{N\times D_0}$. For the decoder part, we use progressive upsampling (PUP) method, as in  SETR \cite{zheng2020rethinking}. Specifically, we first reshape $\mathbf{z}^L$ back to $\mathbf{t}^0\in\mathbb{R}^{\frac{H}{16}\times\frac{W}{16}\times D_0}$, which could be viewed as a 2D feature map with $D_0$ channels. We then use two consecutive standard upsampling-convolution layers to recover the spatial resolution, where we obtain $\mathbf{t}^1\in\mathbb{R}^{\frac{H}{8}\times\frac{W}{8}\times D_1}$ and $\mathbf{t}^2\in\mathbb{R}^{\frac{H}{4}\times\frac{W}{4}\times D_2}$, respectively. The feature maps of different scales $\mathbf{t}^0$, $\mathbf{t}^1$ and $\mathbf{t}^2$ are saved for late fusion with corresponding feature maps of the CNN branch. 

\textit{\textbf{CNN Branch.}}
Traditionally, features are progressively downsampled to $\frac{H}{32}\times \frac{W}{32}$ and hundreds of layers are employed in deep CNNs to obtain global context of features, which results in very deep models draining out resources. Considering the benefits brought by Transformers, we remove the last block from the original CNNs pipeline and take advantage of the Transformer branch to obtain global context information instead. This gives us not only a shallower model but also retaining richer local information. For example, ResNet-based models typically have five blocks, each of which downsamples the feature maps by a factor of two. We take the outputs from the 4th ($\mathbf{g}^0\in\mathbb{R}^{\frac{H}{16}\times \frac{W}{16}\times C_0}$), 3rd ($\mathbf{g}^1\in\mathbb{R}^{\frac{H}{8}\times \frac{W}{8}\times C_1}$) and 2nd ($\mathbf{g}^2\in\mathbb{R}^{\frac{H}{4}\times \frac{W}{4}\times C_2}$) blocks to fuse with the results from Transformer (Fig.~\ref{fig1}).  Moreover, our CNN branch is flexible that any off-the-shelf convolutional network can be applied.  

\textit{\textbf{BiFusion Module.}}
To effectively combine the encoded features from CNNs and Transformers, we propose a new BiFusion module (refer to Fig. \ref{fig1}) that incorporates both self-attention and multi-modal fusion mechanisms. Specifically, we obtain the fused feature representation $\mathbf{f}^{i}, i=0,1,2$ by the following operations:
\begin{equation}
\begin{aligned}
    \mathbf{\hat{t}}^{i} &= \textup{ChannelAttn}(\mathbf{t}^{i})\\ 
     \mathbf{\hat{b}}^{i} &= \textup{Conv}(\mathbf{t}^{i}\mathbf{W}_1^i\odot\mathbf{g}^{i}\mathbf{W}_2^i)
\end{aligned}
\quad \quad
\begin{aligned}
    \mathbf{\hat{g}}^{i} &= \textup{SpatialAttn}(\mathbf{g}^{i})
    \\
    \mathbf{f}^{i} &= \textup{Residual}([\mathbf{\hat{b}}^{i}, \mathbf{\hat{t}}^{i}, \mathbf{\hat{g}}^{i} ])
\end{aligned}
\end{equation}
where $W_1^i\in\mathbb{R}^{D_i\times L_i}$, $W_2^i\in\mathbb{R}^{C_i\times L_i}$, $|\odot|$ is the Hadamard product and Conv is a 3x3 convolution layer. The channel attention is implemented as SE-Block proposed in \cite{hu2018squeeze} to promote global information from the Transformer branch. The spatial attention is adopted from CBAM~\cite{woo2018cbam} block as spatial filters to enhance local details and suppress irrelevant regions, as low-level CNN features could be noisy. The Hadamard product then models the fine-grained interaction between features from 
the two branches. Finally, the interaction features $\mathbf{\hat{b}}^{i}$ and attended features $\mathbf{\hat{t}}^{i}, \mathbf{\hat{g}}^{i}$ are concatenated and passed through a Residual block. The resulting feature $\mathbf{f}^{i}$ effectively captures both the global and local context for the current spatial resolution. To generate final segmentation, $\mathbf{f}^{i}$s are combined using the attention-gated (AG) skip-connection~\cite{schlemper2019attention}, where we have $\mathbf{\hat{f}}^{i+1} =  \textup{Conv}([\textup{Up}(\mathbf{\hat{f}}^{i}), \textup{AG}(\mathbf{f}^{i+1}, \textup{Up}(\mathbf{\hat{f}}^{i}))])$ and $\mathbf{\hat{f}}^{0} =\mathbf{f}^{0} $, as in Fig. \ref{fig1}.

\textit{\textbf{Loss Function.}}
The full network is trained end-to-end with the weighted IoU loss and binary cross entropy loss $L=L_{IoU}^{w} + L_{bce}^{w}$, where boundary pixels receive larger weights \cite{qin2019basnet}. Segmentation prediction is generated by a simple head, which directly resizes the input feature maps to the original resolution and applies convolution layers to generate $M$ maps, where $M$ is the number of classes. Following \cite{fan2020pranet}, We use deep supervision to improve the gradient flow by additionally supervising the transformer branch and the first fusion branch. The final training loss is given by $\mathcal{L} = \alpha L\left(G, \textup{head}(\mathbf{\hat{f}}^2)\right) + \gamma L\left(G, \textup{head}(\mathbf{t}^2)\right)+ \beta L\left(G, \textup{head}(\mathbf{f}^0)\right)$, where $\alpha$, $\gamma$,  $\beta$ are tunnable hyperparameters and $G$ is groundtruth.

\section{Experiments and Results}
\subsubsection{Data Acquisition.}To better evaluate the effectiveness of TransFuse, four segmentation tasks with different imaging modalities, disease types, target objects, target sizes, etc. are considered: 1) \textit{\underline{Polyp Segmentation}}, where five public polyp datasets are used: Kvasir~\cite{jha2020kvasir}, CVC-ClinicDB~\cite{bernal2015wm}, CVC-ColonDB~\cite{tajbakhsh2015automated},
EndoScene~\cite{vazquez2017benchmark} and ETIS~\cite{silva2014toward}. The same split and training setting as described in~\cite{fan2020pranet,huang2021hardnet} are adopted, i.e. 1450 training images are solely selected from Kvasir and CVC-ClinicDB while 798 testing images are from all five datasets. Before processing, the resolution of each image is resized into 352$\times$352 as~\cite{fan2020pranet,huang2021hardnet}. 2) \textit{\underline{Skin Lesion Segmentation}}, where the publicly available 2017 International Skin Imaging Collaboration skin lesion segmentation dataset (ISIC2017)~\cite{codella2018skin} is used\footnote{Another similar dataset ISIC2018 was not used because of the missing test set annotation, which makes fair comparison between existing works can be hardly achieved.
}. ISIC2017 provides 2000 images for training, 150 images for validation and 600 images for testing. 
Following the setting in~\cite{al2018skin}, we resize all images to 192$\times$256.  3) \textit{\underline{Hip Segmentation}}, where a total of 641 cases are collected from a hospital with average size of 2942$\times$2449 and pixel spacing as 0.143mm\footnote{All data are from different patients and with ethics approval, which consists of 267 patients of Avascular Necrosis, 182 patients of Osteoarthritis, 71 patients of Femur Neck Fracture, 33 patients of Pelvis Fracture, 26 patients of Developmental Dysplasia of the Hip and 62 patients of other dieases.}. Each image is annotated by a clinical expert and double-blind reviewed by two specialists. We resized all images into $352\times352$, and randomly split images with a ratio of 7:1:2 for training, validation and testing. 4)\textit{\underline{Prostate Segmentation}}, where volumetric Prostate Multi-modality MRIs from the Medical Segmentation Decathlon \cite{simpson2019large} are used. The dataset contains multi-modal MRIs from 32 patients, with a median volume shape of $20\times320\times319$. Following the setting in~\cite{isensee2019automated}, we reshape all MRI slices to $320\times320$, and independently normalize each volume using z-score normalization.

\subsubsection{Implementation Details.}
TransFuse was built in PyTorch framework~\cite{paszke2019pytorch} and trained using a single NVIDIA-A100 GPU. 
The values of $\alpha$, $\beta$ and $\gamma$ were set to 0.5, 0.3, 0.2 empirically. Adam optimizer with learning rate of 1e-4 was adopted and all models were trained for 30 epochs as well as batch size of 16, unless otherwise specified.

In polyp segmentation experiments, no data augmentation was used except for multi-scale training, as in~\cite{fan2020pranet,huang2021hardnet}.
For skin lesion and hip segmentation, data augmentation including random rotation, horizontal flip, color jittering, etc. were applied during training. A smaller learning rate of 7e-5 was found useful for skin lesion segmentation. 
Finally, we follow the nnU-Net framework~\cite{isensee2019automated} to train and evaluate our model on Prostate Segmentation, using the same data augmentation and post-processing scheme.  
As selected pretrained datasets and branch backbones may affect the performance differently, three variants of TransFuse are provided to 1) better demonstrate the effectiveness as well as flexibility of our approach; 2) conduct fair comparisons with other methods. 
\textit{TransFuse-S} is implemented with ResNet-34 (R34) and 8-layer DeiT-Small (DeiT-S)~\cite{touvron2020training} as backbones of the CNN branch and Transformer branch respectively. Similarly, \textit{TransFuse-L} is built based on Res2Net-50 and 10-layer DeiT-Base (DeiT-B), while  \textit{TransFuse-L*} uses ResNetV2-50 and ViT-B \cite{dosovitskiy2020image}. Note that ViTs and DeiTs have the same backbone architecture and they mainly differ in the pre-trained strategy and dataset: the former is trained on ImageNet21k while the latter is trained on ImageNet1k with heavier data augmentation.  

\subsubsection{Evaluation Results}
\begin{figure}[!h]
\captionsetup{justification=centering}
\begin{center}
\begin{tabular}{>{\centering\bfseries}m{0.72in}>{\centering}m{0.72in}>{\centering}m{0.72in}>{\centering}m{0.72in}>{\centering}m{0.72in}>{\centering\arraybackslash}m{0.72in}}
\multicolumn{6}{c}{\small \textbf{Polyp Segmentation}}\\
\includegraphics[width=0.149\textwidth, height=0.12\textwidth]{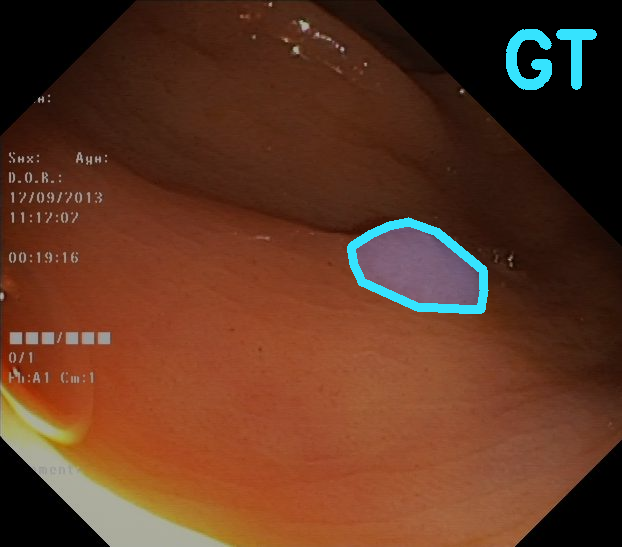}
& \includegraphics[width=0.149\textwidth, height=0.12\textwidth]{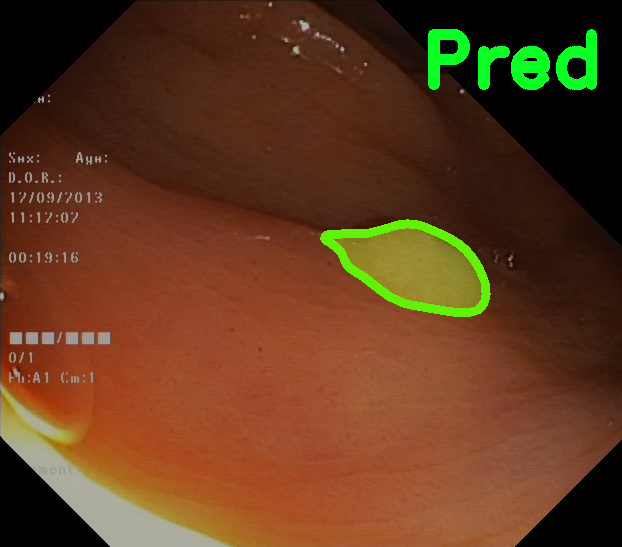}
& \includegraphics[width=0.149\textwidth, height=0.12\textwidth]{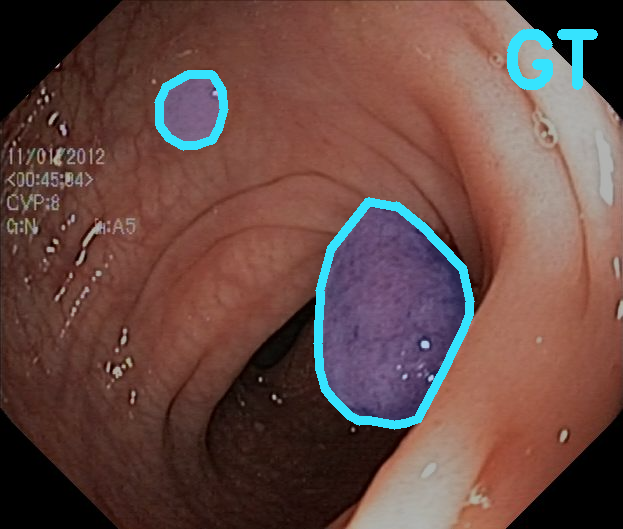}
& \includegraphics[width=0.149\textwidth, height=0.12\textwidth]{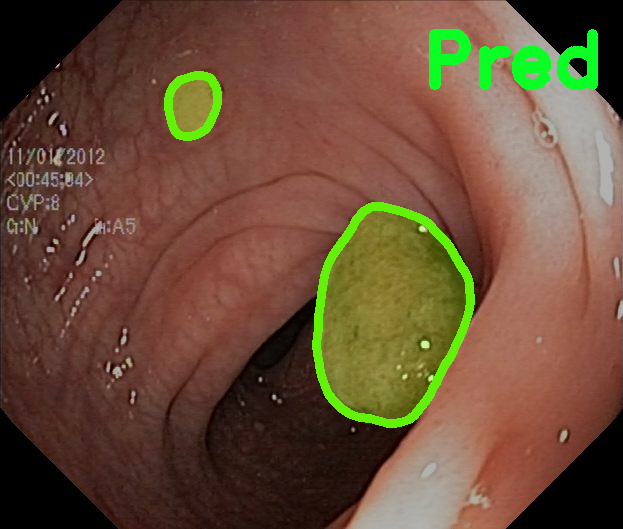}
& \includegraphics[width=0.149\textwidth, height=0.12\textwidth]{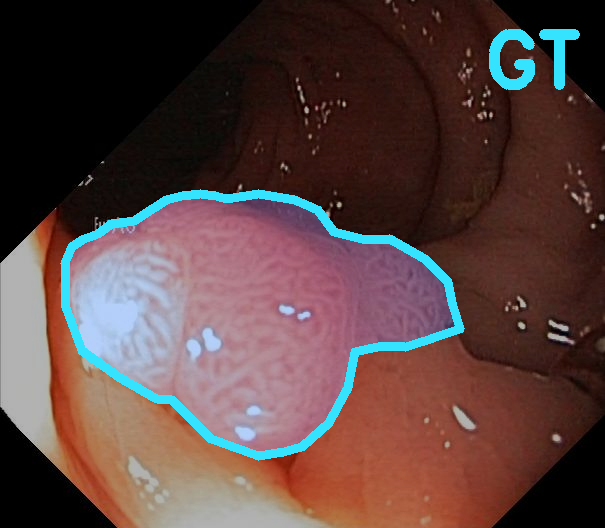}
& \includegraphics[width=0.149\textwidth, height=0.12\textwidth]{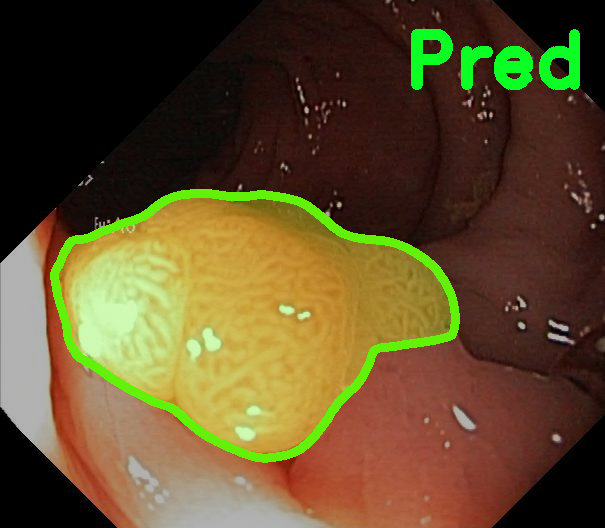}
\\
\multicolumn{6}{c}{\small \textbf{Skin Lesion Segmentation}}\\
\includegraphics[width=0.148\textwidth, height=0.1\textwidth]{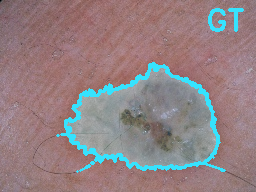}
& \includegraphics[width=0.149\textwidth, height=0.1\textwidth]{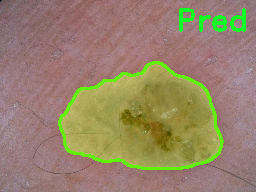}
& \includegraphics[width=0.149\textwidth, height=0.1\textwidth]{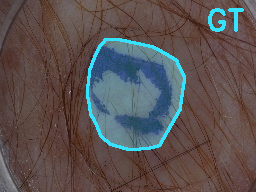}
& \includegraphics[width=0.149\textwidth, height=0.1\textwidth]{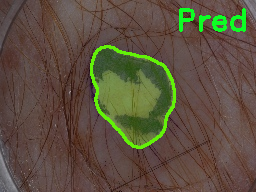}
& \includegraphics[width=0.149\textwidth, height=0.1\textwidth]{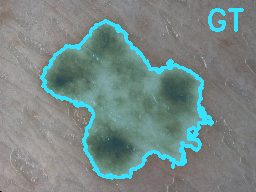}
& \includegraphics[width=0.149\textwidth, height=0.1\textwidth]{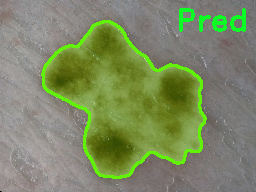}
\end{tabular}
\begin{tabular}{>{\centering\bfseries}m{1.1in}>{\centering}m{1.1in}>{\centering}m{1.1in}>{\centering\arraybackslash}m{1.1in}}
\multicolumn{4}{c}{\small \textbf{Hip Segmentation}}\\
\includegraphics[width=0.23\textwidth, height=0.19\textwidth]{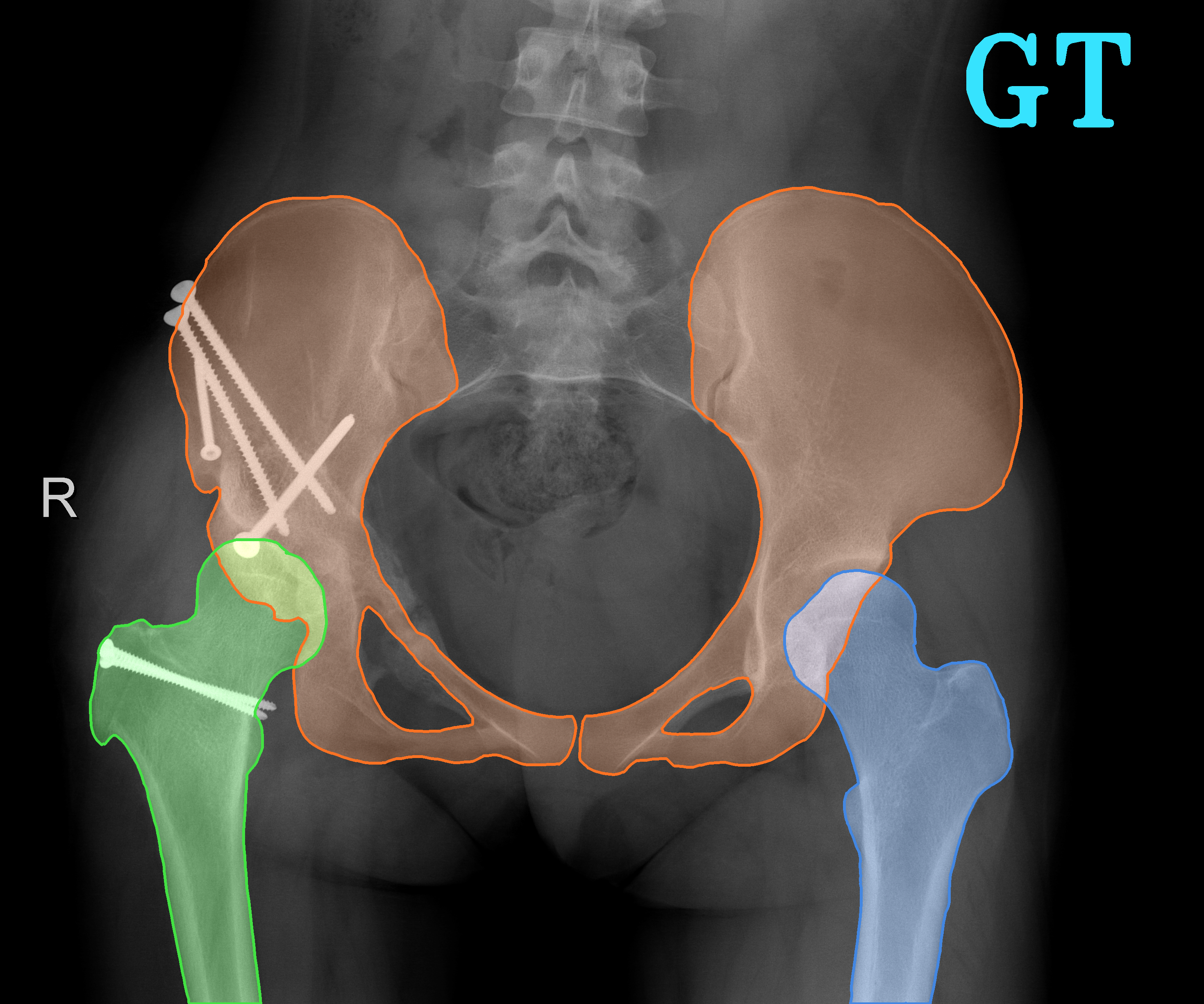}
& \includegraphics[width=0.23\textwidth, height=0.19\textwidth]{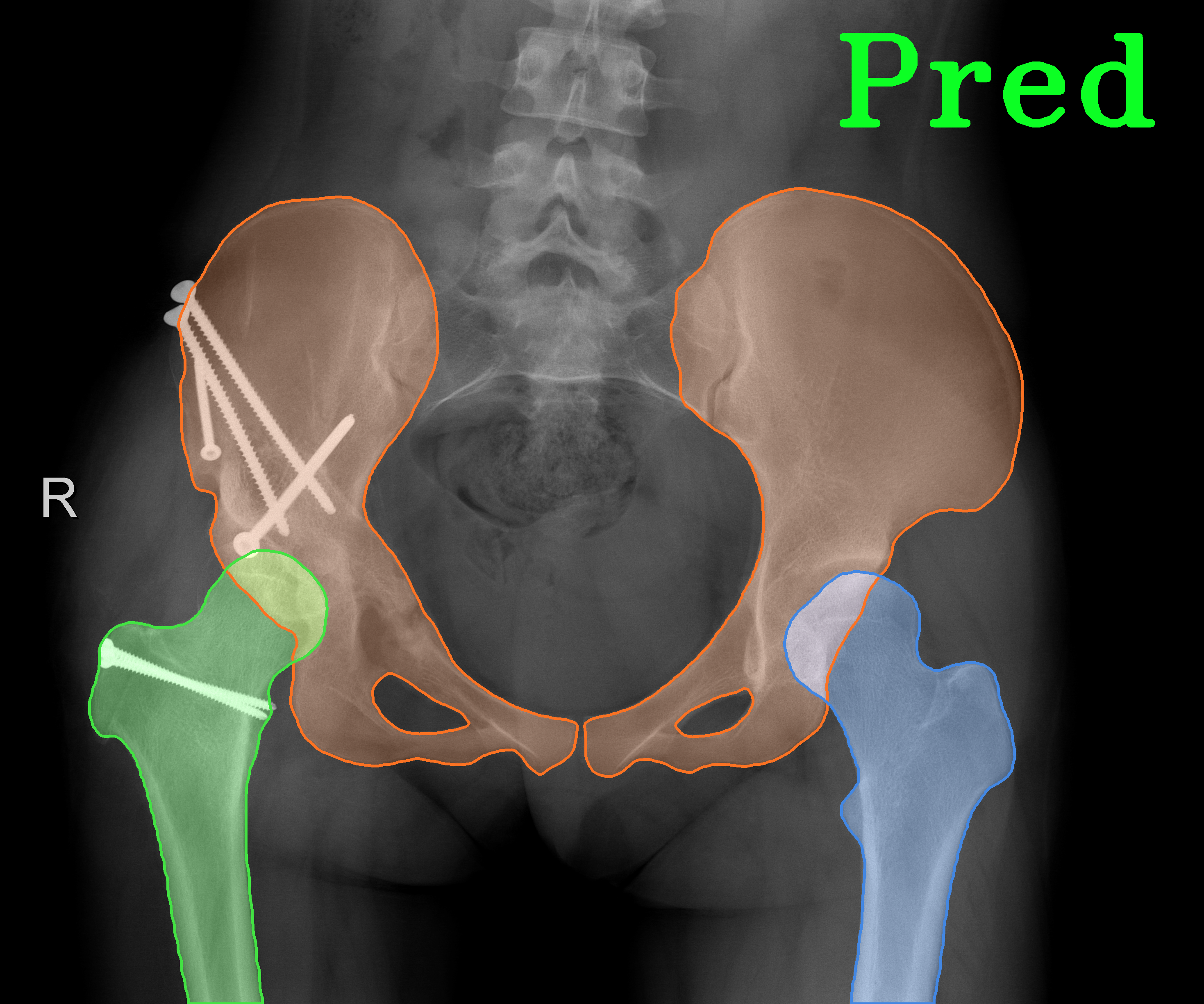}
& \includegraphics[width=0.23\textwidth, height=0.19\textwidth]{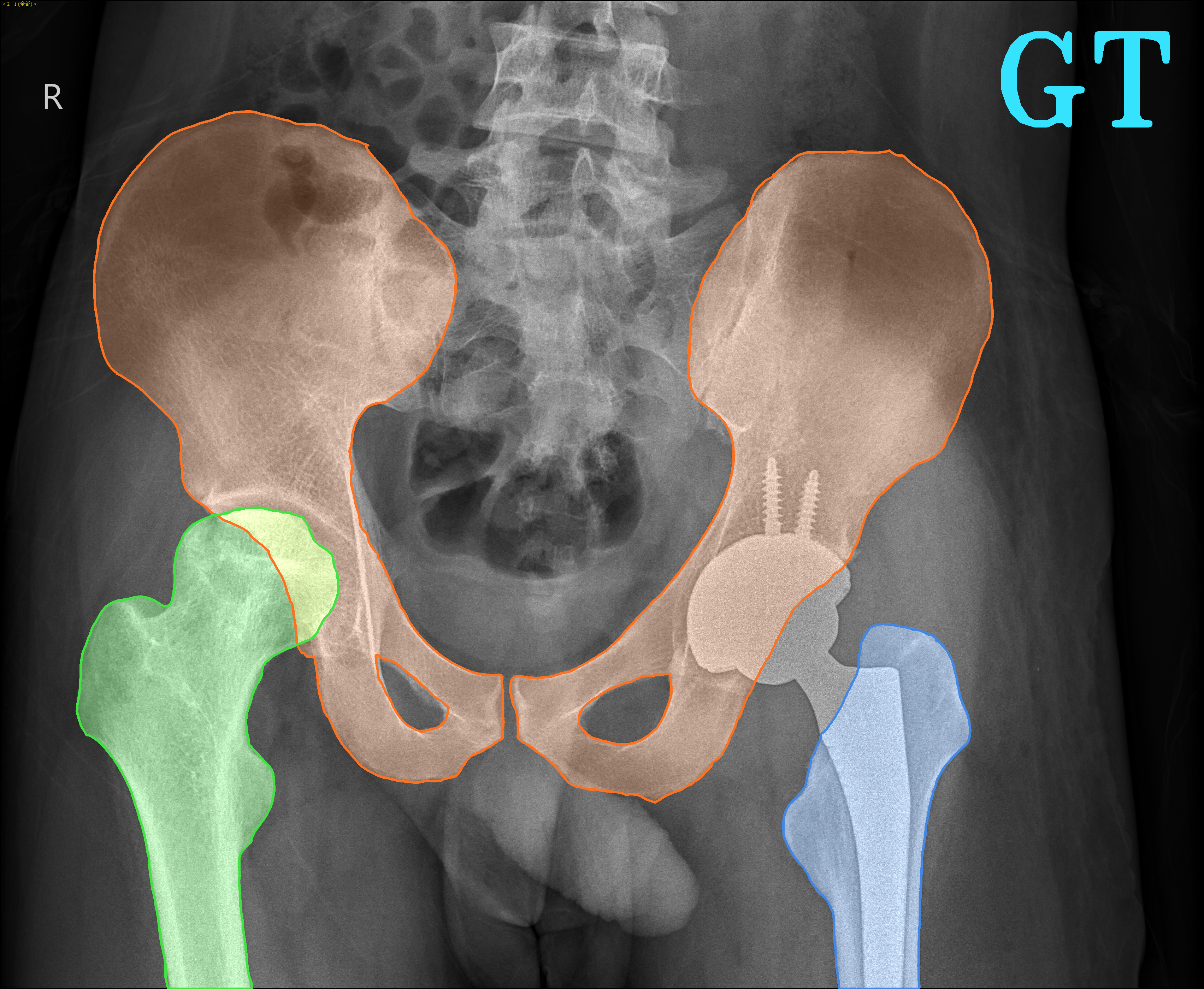}
& \includegraphics[width=0.23\textwidth, height=0.19\textwidth]{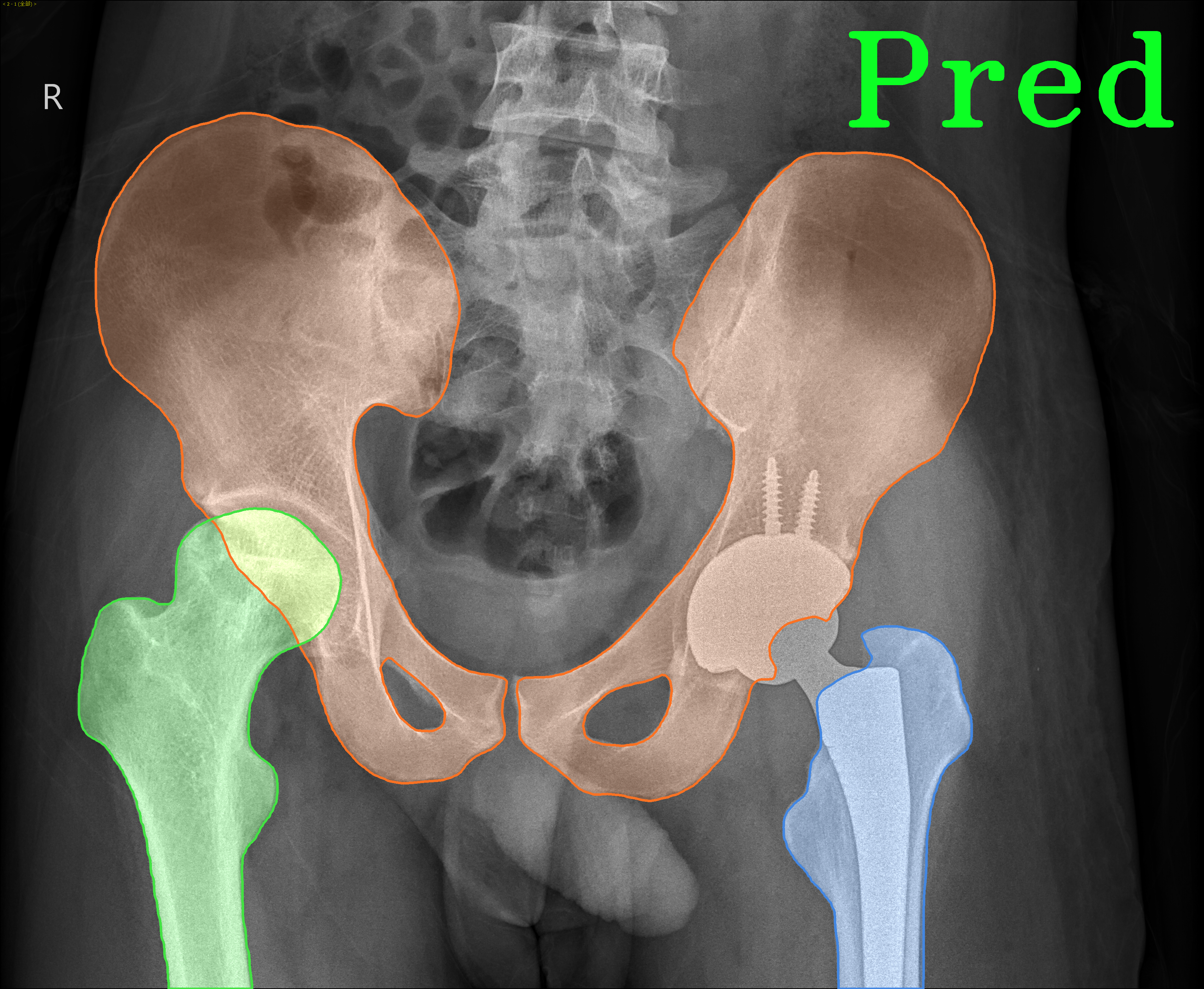}
\end{tabular}
\begin{tabular}{>{\centering\bfseries}m{1.1in}>{\centering}m{1.1in}>{\centering}m{1.1in}>{\centering\arraybackslash}m{1.1in}}
\multicolumn{4}{c}{\small \textbf{Prostate Segmentation}}\\
\includegraphics[width=0.23\textwidth, height=0.19\textwidth]{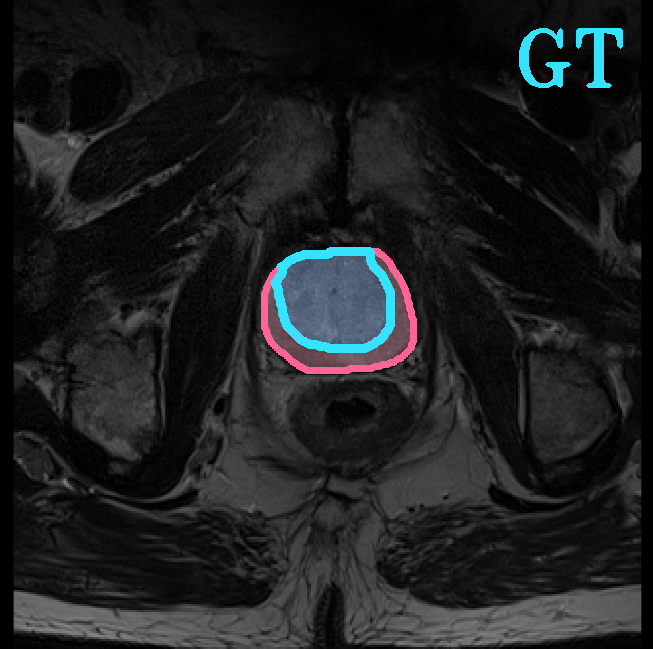}
& \includegraphics[width=0.23\textwidth, height=0.19\textwidth]{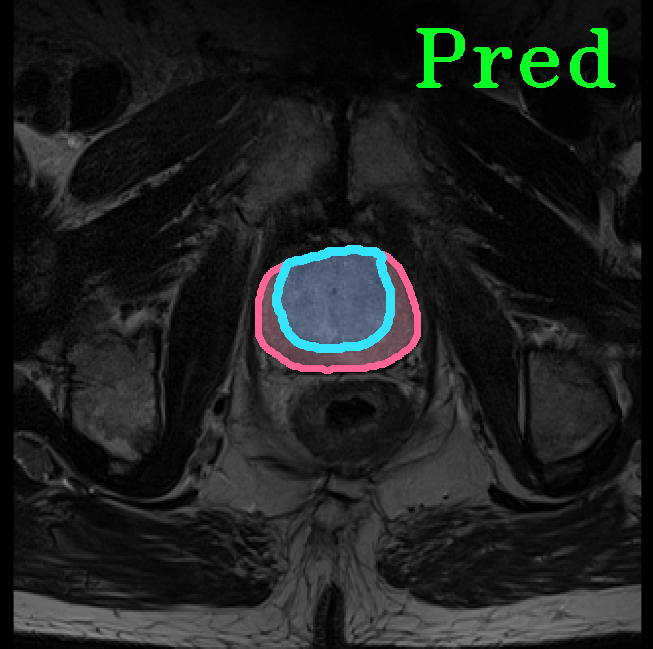}
& \includegraphics[width=0.23\textwidth, height=0.19\textwidth]{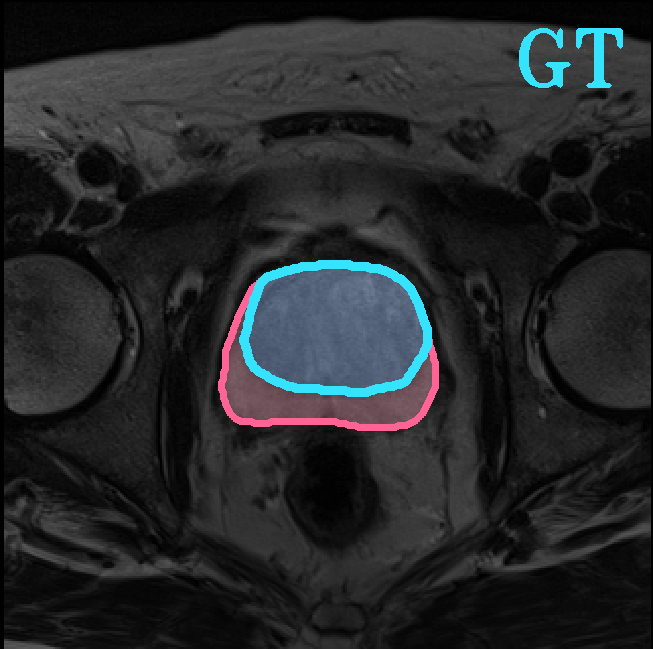}
& \includegraphics[width=0.23\textwidth, height=0.19\textwidth]{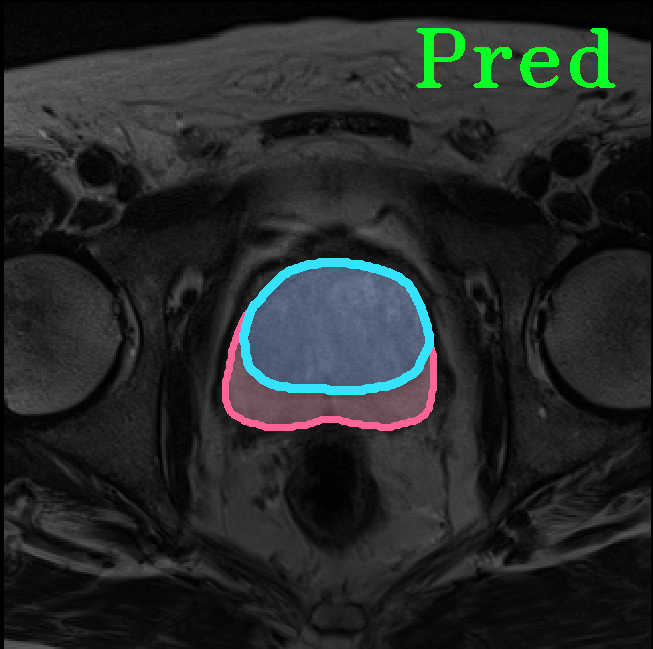}
\end{tabular}

\caption{Results visualization on all three tasks (best viewed in color). Each row follows the repeating sequence of ground truth (GT) and predictions (Pred).}
\label{fig:visResults}
\end{center}
\end{figure}

TransFuse is evaluated on both 2D and 3D datasets to demonstrate the effectiveness. As different medical image segmentation tasks serve different diagnosis or operative purposes, we follow the commonly used evaluation metrics for each of the segmentation tasks to quantitatively analyze the results. Selected visualization results of \textit{TransFuse-S} are shown in Fig.~\ref{fig:visResults}.

\textit{\textbf{Results of Polyp Segmentation.}}
We first evaluate the performance of our proposed method on polyp segmentation against a variety of SOTA methods, in terms of mean Dice (mDice) and mean Intersection-Over-Union (mIoU). 
As in Tab.~\ref{tab:tab1}, our \textit{TransFuse-S/L} outperform CNN-based SOTA methods by a large margin. Specifically, TransFuse-S achieves 5.2\% average mDice improvement on the \textit{unseen} datasets (ColonDB, EndoSene and ETIS). Comparing to other transformer-based methods, \textit{TransFuse-L*} also shows superior learning ability on Kvasir and ClinicDB, observing an increase of 1.3\% in mIoU compared to TransUnet. 
Besides, the efficiency in terms of the number of parameters as well as inference speed is evaluated on an RTX2080Ti with Xeon(R) Gold 5218 CPU. Comparing to prior CNN-based arts, \textit{TransFuse-S} achieves the best performance while using only 26.3M parameters, about 20\% reduction with respect to HarDNet-MSEG (33.3M) and PraNet (32.5M). Moreover, \textit{TransFuse-S} is able to run at 98.7 FPS, much faster than HarDNet-MSEG (85.3 FPS) and PraNet (63.4 FPS), thanks to our proposed parallel-in-branch design. Similarly, \textit{TransFuse-L*} not only achieves the best results compared to other Transformer-based methods, but also runs at 45.3 FPS, about 12\% faster than TransUnet.

\begin{table}[t]
\small
\caption{\label{tab:tab1} Quantitative results on polyp segmentation datasets compared to previous SOTAs. The results of \cite{chen2021transunet} is obtained by running the released code and we implement SETR-PUP. `-' means results not available.} 
\centering
\resizebox{\linewidth}{!}{ 
\begin{tabular}{@{}lcccccccccc@{}}
\toprule
\multirow{2}{*}{Methods} & \multicolumn{2}{c}{Kvasir} & \multicolumn{2}{c}{ClinicDB} & \multicolumn{2}{c}{ColonDB} & \multicolumn{2}{c}{EndoScene} & \multicolumn{2}{c}{ETIS} \\
\multicolumn{1}{c}{} & mDice & mIoU & mDice & mIoU & mDice & mIoU & mDice & mIoU & mDice & mIoU \\ \midrule
U-Net~\cite{ronneberger2015u} & 0.818 & 0.746 & 0.823 & 0.750 & 0.512 & 0.444 & 0.710 & 0.627 & 0.398 & 0.335\\
U-Net++~\cite{zhou2018unet++} & 0.821 & 0.743 & 0.794 & 0.729 & 0.483 & 0.410 & 0.707 & 0.624 & 0.401 & 0.344          \\

ResUNet++~\cite{jha2019resunet++} & 0.813 & 0.793 & 0.796 & 0.796 & - & - & - & - & - & -\\

PraNet~\cite{fan2020pranet} & 0.898          & 0.840          & 0.899          & 0.849          & 0.709          & 0.640          & 0.871          & 0.797          & 0.628          & 0.567          \\
HarDNet-MSEG~\cite{huang2021hardnet} & 0.912          & 0.857          & 0.932 & 0.882 & 0.731          & 0.660          & 0.887          & 0.821          & 0.677          & 0.613          \\

\textit{TransFuse-S} & \textbf{0.918} & \textbf{0.868} & 0.918          & 0.868          & \textbf{0.773} & \textbf{0.696} & 0.902 & 0.833 & 0.733 & 0.659 \\
\textit{TransFuse-L} & \textbf{0.918} & \textbf{0.868} & \textbf{0.934}          & \textbf{0.886}          & 0.744 & 0.676 & \textbf{0.904} & \textbf{0.838} & \textbf{0.737} & \textbf{0.661} \\
\hline
SETR-PUP~\cite{zheng2020rethinking} & 0.911 & 0.854 & 0.934 & 0.885 & 0.773 & 0.690 & 0.889 & 0.814 & 0.726 & 0.646 \\
TransUnet~\cite{chen2021transunet} & 0.913 & 0.857 & 0.935 & 0.887  & 0.781 & 0.699 & 0.893 & 0.824 & 0.731 & 0.660 \\
\textit{TransFuse-L*} & \textbf{0.920} & \textbf{0.870} & \textbf{0.942}          & \textbf{0.897}          & \textbf{0.781} & \textbf{0.706} & \textbf{0.894} & \textbf{0.826} & \textbf{0.737} & \textbf{0.663} \\
\bottomrule
\end{tabular}}

\parbox{.52\textwidth}{
\centering
\caption{ Quantitative results on ISIC 2017 test set. Results with backbones use weights pretrained on ImageNet.}\label{tab:skin}
\resizebox{0.53\textwidth}{!}{
\begin{tabular}{lccccc}
\toprule
Methods      & Backbones & Epochs  & Jaccard & Dice & Accuracy \\ \midrule
CDNN~\cite{yuan2017improving} &  - & -  & 0.765 & 0.849 & 0.934\\
DDN~\cite{li2018dense}      & ResNet-18  & 600         & 0.765     & 0.866 & 0.939                       \\
FrCN~\cite{al2018skin} & VGG16   & 200  & 0.771  & 0.871  & 0.940\\
DCL-PSI~\cite{bi2019step} & ResNet-101 & 150  & 0.777 & 0.857 & 0.941  \\
SLSDeep~\cite{sarker2018slsdeep}     & ResNet-50        & 100       & 0.782 & \textbf{0.878}     & 0.936                  \\
\midrule
Unet++~\cite{zhou2018unet++} & ResNet-34 & 30 & 0.775  & 0.858 & 0.938  \\
\textit{TransFuse-S} & R34+DeiT-S        & 30             & \textbf{0.795}     & 0.872  & \textbf{0.944}                     \\ \bottomrule
\end{tabular}}
}
\hfill
\parbox{.44\textwidth}{
\centering
\caption{\label{tab:hip} Results on in-house hip dataset. All models use pretrained backbones from ImageNet and are of similar size ($\sim$ 26M). HD and ASD are measured in mm.}
\resizebox{0.44\textwidth}{!}{
\begin{tabular}{@{}lccccccc@{}}
\toprule
\multirow{2}{*}{Methods}  & \multicolumn{2}{c}{Pelvis} & \multicolumn{2}{c}{L-Femur} & \multicolumn{2}{c}{R-Femur}\\
\multicolumn{1}{c}{} & HD & ASD & HD & ASD & HD & ASD\\ \midrule

Unet++~\cite{zhou2018unet++} &  14.4 & 1.21 & 9.33 & 0.932 & 5.04 & 0.813\\
HRNetV2~\cite{wang2020deep} &  14.2 & 1.13 & 6.36 & 0.769 & 5.98 & 0.762\\
\textit{TransFuse-S} & \textbf{9.81} & \textbf{1.09} & \textbf{4.44} & \textbf{0.767} & \textbf{4.19} & \textbf{0.676}\\ \bottomrule
\end{tabular}}}
\end{table}

\textit{\textbf{Results of Skin Lesion Segmentation.}}
The ISBI 2017 challenge ranked methods according to Jaccard Index~\cite{codella2018skin} on the ISIC 2017 test set. Here, we use Jaccard Index, Dice score and pixel-wise accuracy as evaluation metrics. The comparison results against leading methods are presented in Tab.~\ref{tab:skin}. \textit{TransFuse-S} is about 1.7\% better than the previous SOTA SLSDeep~\cite{sarker2018slsdeep} in Jaccard score, without any pre- or post-processing and converges in less than $1/3$ epochs. Besides, our results outperform Unet++~\cite{zhou2018unet++} that employs pretrained R34 as backbone and has comparable number of parameters with TransFuse-S (26.1M vs 26.3M). Again, the results prove the superiority of our proposed architecture.

\textit{\textbf{Results of Hip Segmentation.}}
Tab.~\ref{tab:hip} shows our results on hip segmentation task, which involves three human body parts: Pelvis, Left Femur (L-Femur) and Right Femur (R-Femur). Since the contour is more important in dianosis and THA preoperative planning, we use Hausdorff Distance (HD) and Average Surface Distance (ASD) to evaluate the prediction quality. Compared to the two advanced segmentation methods \cite{zhou2018unet++,wang2020deep}, \textit{TransFuse-S} performs the best on both metrics and reduces HD significantly (30\% compared to HRNetV2 as well as 34\% compared to Unet++ on average), indicating that our proposed method is able to capture finer structure and generates more precise contour.

\textit{\textbf{Results of Prostate Segmentation.}}
We compare TransFuse-S with nnU-Net~\cite{isensee2019automated}, which ranked 1st in the prostate segmentation challenge~\cite{simpson2019large}. We follow the same preprocessing, training as well as evaluation schemes of the publicly available nnU-Net framework\footnote{https://github.com/MIC-DKFZ/nnUNet} and report the 5-fold cross validation results in Tab.~\ref{tab:prostate}. We can find that TransFuse-S surpasses nnUNet-2d by a large margin (+4.2\%) in terms of the mean dice score. Compared to nnUNet-3d, TransFuse-S not only achieves better performance, but also reduces the number of parameters by $\sim$41\% and increases the throughput by $\sim$50\% (on GTX1080).

\textit{\textbf{Ablation Study.}} An ablation study is conducted to evaluate the effectiveness of the parallel-in-branch design as well as BiFusion module by varying design choices of different backbones, compositions and fusion schemes. A \textit{seen} (Kvasir) and an \textit{unseen} (ColonDB) datasets from polyp are used, and results are recorded in mean Dice. In Tab.~\ref{tab:alb}, by comparing E.3 against E.1 and E.2, we can see that combining CNN and Transformer leads to better performance. Further, by comparing E.3 against E.5, E.6, we observe that the parallel models perform better than the sequential counterpart. Moreover, we evaluate the performance of a double branch CNN model (E.4) using the same parallel structure and fusion settings with our proposed E.6. We observe that E.6 outperforms E.4 by 2.2\% in Kvasir and 18.7\% in ColonDB, suggesting that the CNN branch and transformer branch are complementary to each other, leading to better fusion results. 
Lastly, performance comparison is conducted between another fusion module comprising concatenation followed by a residual block and our proposed BiFusion module (E.5 and E.6). Given the same backbone and composition setting, E.6 with BiFusion achieves better results. Additional experiments conducted on ISIC2017 are presented in Tab.~\ref{tab:bifusion} to verify the design choice of BiFusion module, from which we find that each component shows its unique benefit.

\begin{table}[t]
\small
\caption{\label{tab:prostate} Quantitative results on prostate MRI segmentation. PZ, TZ stand for the two labeled classes (peripheral and transition zone) and performance (PZ, TZ and mean) is measure by dice score.}
\centering
\resizebox{0.8\linewidth}{!}{
\begin{tabular}{lccccc}
\toprule
Methods      & PZ & TZ  & Mean & Params & Throughput  \\ 
\midrule
nnUnet-2d~\cite{isensee2019automated} & 0.6285  & 0.8380  & 0.7333 & 29.97M & 0.209s/vol\\
nnUnet-3d$\_$full\cite{isensee2019automated}      & 0.6663  & 0.8410         & 0.7537     & 44.80M & 0.381s/vol                       \\
\midrule
TransFuse-S & \textbf{0.6738}        & \textbf{0.8539}             & \textbf{0.7639}     & \textbf{26.30M}  & \textbf{0.192s/vol}                     \\ \bottomrule
\end{tabular}}
\parbox{.54\textwidth}{
\centering
\caption{\label{tab:alb} Ablation study on parallel-in-branch design. Res: Residual.}
\resizebox{\linewidth}{!}{
\begin{tabular}{@{}lccccc@{}}
\toprule
Index & Backbones & Composition  &  Fusion    & Kvasir & ColonDB \\ \midrule

E.1 & R34 & Sequential & - & 0.890 & 0.645 \\
E.2 & DeiT-S & Sequential & - & 0.889 & 0.727 \\
E.3 & R34+DeiT-S & Sequential & - & 0.908  & 0.749   \\
E.4 & R34+VGG16  & Parallel   & BiFusion    & 0.896  & 0.651   \\

E.5 & R34+DeiT-S & Parallel   & Concat+Res      & 0.912  & 0.764   \\
E.6 & R34+DeiT-S & Parallel   & BiFusion    & 0.918  & 0.773   \\ \bottomrule
\end{tabular}}
}
\hfill
\parbox{.43\textwidth}{
\centering
\caption{\label{tab:bifusion} Ablation study on BiFusion module. Res: Residual; TFM: Transformer; Attn: Attention.}
\resizebox{\linewidth}{!}{
\begin{tabular}{lccccc}
\toprule
Fusion & Jaccard & Dice  & Accuracy  \\ 
\midrule
Concat+Res & 0.778 & 0.857 & 0.939 \\
+CNN Spatial Attn & 0.782 & 0.861 & 0.941\\
+TFM Channel Attn & 0.787 & 0.865 & 0.942\\
+Dot Product & 0.795 & 0.872 & 0.944\\
 \bottomrule
\end{tabular}}
}
\end{table}

\section{Conclusion}
In this paper, we present a novel strategy to combine Transformers and CNNs with late fusion for medical image segmentation. The resulting architecture, TransFuse, leverages the inductive bias of CNNs on modeling spatial correlation and the powerful capability of Transformers on modelling global relationship. TransFuse achieves SOTA performance on a variety of segmentation tasks whilst being highly efficient on both the parameters and inference speed. We hope that this work can bring a new perspective on using transformer-based architecture. In the future, we plan to improve the efficiency of the vanilla transformer layer as well as test TransFuse on other medical-related tasks such as landmark detection and disease classification.

\subsubsection{Acknowledgement.} 
We gratefully thank Weijun Wang, MD, Zhefeng Chen, MD, Chuan He, MD, Zhengyu Xu, Huaikun Xu for serving as our medical advisors on hip segmentation project. 

\bibliographystyle{splncs04}
\bibliography{mybibliography}

\end{document}